\documentclass[lettersize,journal]{IEEEtran}
\usepackage{amsmath,amsfonts,amssymb}
\usepackage{algorithmic}
\usepackage{algorithm}
\usepackage{array}
\usepackage[caption=false,font=normalsize,labelfont=sf,textfont=sf]{subfig}
\usepackage{textcomp}
\usepackage{stfloats}
\usepackage{url}
\usepackage{verbatim}
\usepackage{graphicx}
\usepackage{cite}
\usepackage{colortbl}

\newcommand{\hlight}[1]{\textcolor{black}{#1}}

\begin{document}

\title{Haptic Guidance and Haptic Error Amplification \\ in a Virtual Surgical Robotic Training Environment}

\author{Yousi A. Oquendo$^1$, Margaret M. Coad$^2$,~\IEEEmembership{Member,~IEEE,} Sherry M. Wren, Thomas S. Lendvay, Ilana Nisky,~\IEEEmembership{Senior Member,~IEEE,} Anthony M. Jarc, Allison M. Okamura,~\IEEEmembership{Fellow, IEEE,} Zonghe Chua$^2$,~\IEEEmembership{Member,~IEEE}
\thanks{Manuscript submitted dd/mm/yyyy}
\thanks{This work was supported by and educational grant from Intuitive Surgical, Inc. and Israeli Science Foundation (Grant 327/20)}
\thanks{Y.A. Oquendo is with the Hospital for Special Surgery, New York, NY}
\thanks{M.M. Coad is with the Department of Aerospace and Mechanical Engineering at the University of Notre Dame, South Bend, IN}
\thanks{S.M. Wren is with the Department of Surgery in the School of Medicine at Stanford University, Stanford, CA}
\thanks{T.S. Lendvay is with the Department of Urology in the School of Medicine at the University of Washington, Seattle,WA}
\thanks{I. Nisky is with the Department of Bioengineering and the School of Brain Sciences and Cognition at Ben-Gurion University of the Negev, Israel}
\thanks{A.M. Jarc is with Intuitive Surgical Inc., Sunnyvale, CA}
\thanks{A.M. Okamura is with the Department of Mechanical Engineering at Stanford University, Stanford, CA}
\thanks{Z. Chua \emph{(corresponding author: zonghe.chua@case.edu)} is with the Department of Electrical, Computer, and Systems Engineering at Case Western Reserve University, Cleveland, OH }
\thanks{$^1$ Work performed while in the School of Medicine at Stanford University}
\thanks{$^2$ Work performed while in the Department of Mechanical Engineering at Stanford University}
}

\markboth{IEEE Transaction on Haptics}%
{Shell \MakeLowercase{\textit{et al.}}: A Sample Article Using IEEEtran.cls for IEEE Journals}


\maketitle

\begin{abstract}
Teleoperated robotic systems have introduced more intuitive control for minimally invasive surgery, but the optimal method for training remains unknown. Recent motor learning studies have demonstrated that exaggeration of errors helps trainees learn to perform tasks with greater speed and accuracy. We hypothesized that training in a force field that pushes the operator away from a desired path would improve their performance on a virtual reality ring-on-wire task.

Forty surgical novices trained under a no-force, guidance, or error-amplifying force field over five days. Completion time, translational and rotational path error, and combined error-time were evaluated under no force field on the final day. The groups significantly differed in combined error-time, with the guidance group performing the worst. Error-amplifying field participants showed the most improvement and did not plateau in their performance during training, suggesting that learning was still ongoing. Guidance field participants had the worst performance on the final day, confirming the guidance hypothesis. Participants with high initial path error benefited more from guidance. Participants with high initial combined error-time benefited more from guidance and error-amplifying force field training. Our results suggest that error-amplifying and error-reducing haptic training for robot-assisted telesurgery benefits trainees of different abilities differently.
\end{abstract}






\section{Introduction}
\label{S:1}
\IEEEPARstart{S}{urgical} procedures require a high level of technical skill to ensure efficiency and protect the safety of the patient. Due to the direct effect of surgeon skill on patient outcomes, there is a need for cost-effective and realistic training methods that both decrease training time and reduce training error \cite{birkmeyer2013surgicaloutcomes}. This need is underscored by the constant development of new instruments and techniques for the operating room. Robot-assisted minimally invasive surgical (RMIS) platforms such as the da Vinci Surgical System (Intuitive Surgical, CA, USA) allow for more intuitive and ergonomic control while adding the challenge of teleoperation with limited haptic feedback \cite{moorthy2004dexterity}. Despite its challenges, the nature of teleoperation as a two-sided system with a surgeon-side console and a physically separate patient-side console offers a unique opportunity to develop training tools and through precision sensing, understand the evolution of surgical skill in robot-assisted minimally invasive surgery.

 VR simulators \cite{Davila2017} in particular offer several advantages to other simulators. Compared to their physical counterparts, VR simulators are more space-efficient, allow for design of custom environments, provide easily replicated simulation experiences across different users, locations and robotic systems, and allow for automated collection of data.  In addition to quantification of movement, automated data collection more broadly allows for assessment of surgical skill and addition of real-time feedback. Further, the ability of RMIS systems to apply forces to the hands of the surgeon and augment visual information provides a promising method of potentially expediting surgeon training \cite{van2009valueofhaptics}.

Despite their many advantages, VR systems have historically been criticized for their lack of tactile feedback and questionable translation to operating room performance \cite{Norkhairani2011}\cite{Singapogu2008}. Recent studies have described methods with a demonstrated translation to surgical skill \cite{youngblood2005VRcomparison} \cite{cho2013virtual}, and improvements in computer processing speed and virtual reality technology have resulted in newer systems with varying implementations of haptic feedback and/or guidance \cite{Overtoom2019}. In the motor learning field, the effects of haptic guidance on learning a task have been inconclusive \cite{Sigrist2013guidanceReview}. The negative results are typically said to support the \textit{Guidance Hypothesis}, which states that assistive feedback results in over-reliance on the feedback for performance, leading to poor results when the feedback is removed \cite{schmidt1989guidancehypothesis}\cite{powell2012sharedcontrol}. This has led to the exploration of the opposite paradigm, that of error amplification through haptic feedback \cite{Sigrist2013guidanceReview}. The results of such studies have generally shown promise for learning especially for more skilled users \cite{milotComparisonErroramplificationHapticguidance2010} and also in rehabilitation settings \cite{pattonSensoryMotorInteractionsError2016}.  

In this study, we aim to build on the work from two previous studies. The first study by Coad et al. \cite{coad2017training} required users to complete a simple 2D path-following task with guidance forces or error amplification forces. In this study, the group that trained with error amplification forces initially showed superior performance compared to the groups that had either trained with no force feedback, or with guidance forces. However, by the end of the evaluation, all groups displayed similar levels of performance. One possible explanation proposed for the lack of differences between training groups was that the task was too simple and had a short learning curve that was overcome quickly by all groups. The second study by Enayati et al.\cite{enayati2018assistance} in contrast required users to complete a more complex 3D VR path-following task but with only guidance forces that adapted to the user's performance. In this study, the adaptive assistance group showed superior performance in time to completion but otherwise similar performance to the unassisted group.

In this investigation, we aim to evaluate the effect of guidance forces and error amplification forces designed in Coad et al.\,\cite{coad2017training} on the performance of the more complex 3D virtual path-following task developed in Enayati et al.\,\cite{enayati2018assistance} We hypothesize that participants who train with error amplification forces will show superior performance to those who either receive no force feedback or those who receive assistance from guidance forces. 

In addition, we hypothesize that users of different baseline skill levels will learn differently under different training conditions. While users are required to complete the same path-following task, the addition of error amplification forces adds an element of difficulty to training for those assigned to this training condition. In contrast, the addition of guidance forces allows for participants to more easily complete the same task. The optimal challenge point framework states that there is an optimal amount of task difficulty for a participant's skill level \cite{Guadagnoli2004}. This idea was previously tested with gradually reduced haptic guidance training by Enayati et al.\,with inconclusive results \cite{enayati2018assistance}. The presentation of three training conditions of varying difficulty to individual participants allows us the opportunity to investigate the predictions of the framework with respect to individuals' baseline skill, and with increased difficulty from haptic error amplification as opposed to varying amounts of haptic guidance. We predict that participants with lower baseline skill will benefit more from haptic guidance training compared to those with higher baseline skill. Accordingly, we predict the opposite trend with respect to baseline skill for those training with haptic error amplification.







\section{Methods}
\label{S:2}

\subsection{Experimental Setup}

\begin{figure}[!t]
\centering\includegraphics[width=\linewidth]{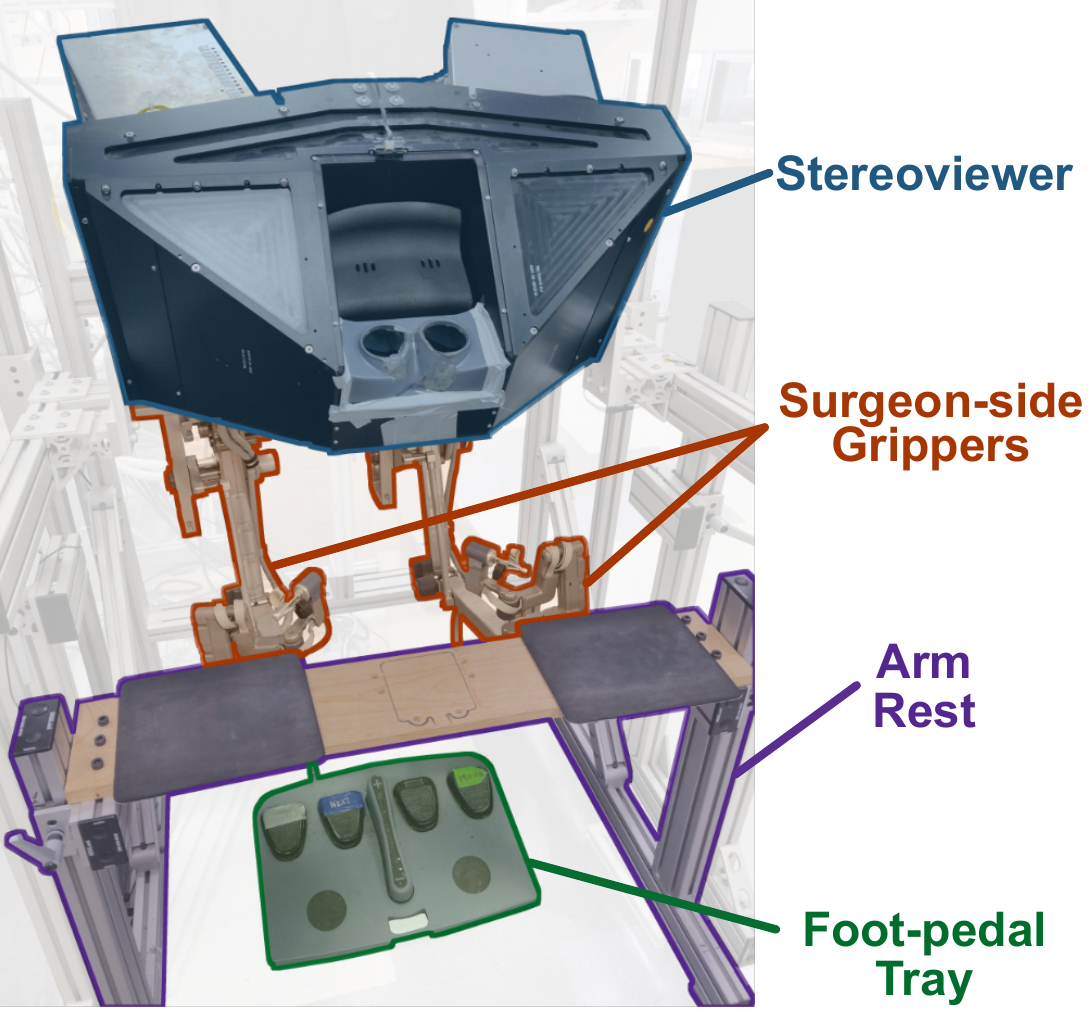}
\caption{The da Vinci Research Kit. The robot was programmed to only allow teleoperation when the rightmost pedal was continuously depressed.}
\label{Console}
\end{figure}

This study was conducted using the surgeon console of a da Vinci Research Kit (dVRK).  The dVRK is a first-generation da Vinci system integrated with open-source control hardware and software, allowing for measurement of movement and sending of commands directly to the joints of each manipulator \cite{kazanzides2014dvrk}. Our study used only the surgeon console of the dVRK, which consisted of a stereoviewer (640$\times$480 resolution), two surgeon-side grippers with seven joints and a gripper each, an arm rest, and a foot-pedal tray (Figure \ref{Console}). The teleoperation was programmed to only be enabled when the participant presses and holds the rightmost pedal (Figure \ref{Console}).
Participants looked into the stereoviewer to see a 3D stereoscopic rendering of the virtual environment.

\begin{figure*}[h]
\centering\includegraphics[width=\linewidth]{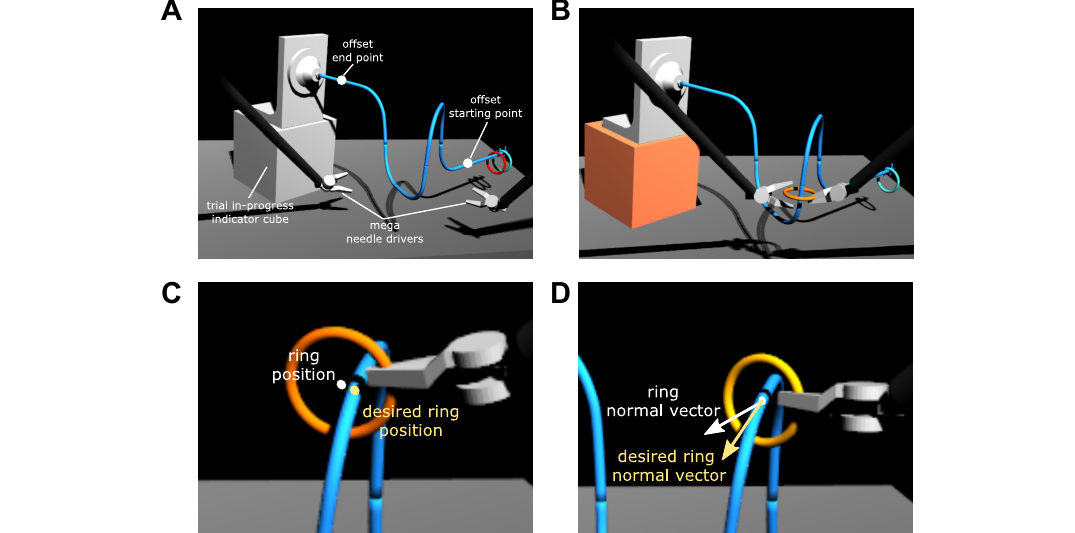}
\caption{(A) The ATAR virtual environment. The indicator cube turns orange once the user crosses the starting point and back to gray when the user crosses the end point. Force feedback is enabled before the starting point. (B) The ring transfer maneuver. The right manipulator is typically in an underhand grip. (C) The current center of the ring (white) and the desired position (yellow). (D) The ring normal vector (white) and the desired normal vector (yellow).}
\label{atarsummary}
\end{figure*}

The VR environment was comprised of a curved wire that traverses all three Cartesian dimensions, two rings, two patient-side mega needle drivers that scaled motion from the surgeon-side gripper by a factor of 0.6, a floor, and a 3D polygonal object to which one end of the wire was attached (Figure\ \ref{atarsummary}A). This environment was adapted from the previously developed Assisted Teleoperation with Augmented Reality (ATAR) framework \cite{enayati2017framework}. 


\subsection{Task}

The visuomotor task used for evaluation was a classic ring-on-wire steady-hand exercise often included in robotic surgery training curricula \cite{smith2014fundamentals}. The virtual rendering of the task was adapted from \cite{enayati2018assistance}. In the ring-on-wire task, participants guided one of the two rings from the starting point of the wire on the right side of the environment to the end point of the wire attached to the gray polygonal object on the left side of the environment. The objective of the task was to move the ring as quickly and accurately as possible from start to finish while keeping the ring in its ideal pose and avoiding contact with the wire. The ideal pose of the ring consist of its ideal position and ideal orientation relative to the wire. The ideal position of the ring was defined as the position that allowed the ring to be centered such that each portion of the ring was equidistant from the wire \mbox{(Figure\ \ref{atarsummary}C)}. The ideal orientation of the ring was defined as the orientation where the circular face of the ring was perpendicular to the wire, such that the vector normal to the plane of the ring was lined up with the tangent of the wire (Figure\ \ref{atarsummary}D). Successful completion of the task required three $90^\circ$ wrist rotations from each hand. Participants were required to grasp the ring from the starting point on the right side of the wire with the right instrument, carry it to the point halfway between the start and end of the wire, transfer the ring to the left instrument (Figure\ \ref{atarsummary}B), and carry the ring to the end point on the left side of the wire.



Participants received visual feedback of their pose (position and orientation) via direct observation of the 3D environment. Position error was further highlighted using color guidance -- the ring changed from yellow to orange to red depending on the distance of the center of the ring from the wire (Figure\,\ref{atarsummary}). The color was continuous from yellow to red and directly correlated to distance, such that a perfectly centered ring was rendered as yellow and a ring that was one ring radius length away from the wire was rendered as red. Participants also received visual feedback on their completion status via color: the cube supporting the virtual wire on the left side of the environment, termed the indicator cube in Figure\ \ref{atarsummary}A, turned from grey to orange when the participant began a trial and subsequently turned from orange to grey upon completion of the task. The starting and ending positions are highlighted in Figure\ \ref{atarsummary}A. Data was collected and recorded at 30 Hz from the starting point to the end point. 


Several alterations were made to the original ATAR VR environment \cite{enayati2017framework} to increase the difficulty of the task and thus require a longer training time to achieve proficiency. First, the friction between the rings and the wire was intentionally reduced such that participants needed to pay close attention to how they gripped the ring. Hitting the ring with the gripper without grasping the ring would result in the ring sliding along the wire. If this occurred prior to the beginning of data recording and the participant hit the ring such that it passed the starting point, recording would begin despite the participant not having grasped the ring. Second, the physics of the wire were altered such that the ring was able to be pulled off the wire. If the participant were to drop the ring while the wire was still contained within the inner diameter of the ring, it would stay on the wire and behave with normal physics. However, if the participant pulled the ring at high velocity, the ring would come off the wire with no resistance. Similarly, if a participant approached the ring hanging on the wire at high velocity, the ring would fall off of the wire. Participants were required to restart a trial if they dropped the ring on the floor of the environment, but were allowed to continue if they dropped the ring onto the wire. Third, the wire exerted a force on the ring if the participant was grasping the ring and dragging it along the wire. This effect encouraged participants to keep the ring centered rather than dragging it along the wire throughout the desired path. Lastly, the two instruments could not grip the ring at the same time. This intentionally made the transfer portion of the task difficult, such that participants were required to balance the ring on one arm of the left instrument gripper and let go of the ring with the gripper of the right instrument before closing the left gripper in order to successfully transfer the ring (Figure\ \ref{atarsummary}B). Ultimately, these changes were implemented to increase the overall difficulty of the task. Specifically, these alterations were implemented to introduce elements to the task that were unfamiliar to the user, forcing participants to adapt to new situations in a manner analogous to performing a surgical procedure for the first several times.

\subsection{Force Field Generation}

Participants who received guidance feedback received an assistive force and torque that pushed and rotated the surgeon-side gripper toward its ideal pose. Those who received error-amplifying feedback experienced an inhibiting force and torque that pushed and rotated the surgeon-side gripper away from its ideal pose. 

For each current pose ($P_C = [T_C, Q_C]$) and desired pose ($P_D = [T_D, Q_D]$), where $T_X$ is a 3-DOF position vector and $Q_X$ is a 4-DOF unit quaternion, the force (F) and  torque ($ \tau $) produced on the gripper were: 
\begin{equation}
\label{eq: forcePD}
F = -k_T(T_D-T_C)-d_T \dot{T_C},
\end{equation}
\begin{equation}
\label{eq: torquePD}
\tau = -k_R \ H_{rpy}(Q_DQ_C^{-1})-d_R \omega_C
\end{equation}
where $\dot{T_C} \textrm{ and } \omega_C$ are the current translational and angular rate, respectively, $k_T$ and $k_R$ are the translational and rotational proportional coefficients, $d_T$ and $d_R$ are the translational and rotational damping coefficients, and $H_{rpy}()$ represents a transformation from quaternion to roll-pitch-yaw. $k_T,\ k_R,\ d_T \textrm{ and } d_R$ were positive in the guidance force field and negative in the error-amplifying force field.




\subsection{Procedure}

\newcolumntype{P}[1]{>{\centering\arraybackslash}p{#1}}
\begin{table}
\centering
\caption{Breakdown of Participant Backgrounds}
\resizebox{\linewidth}{!}{
\begin{tabular}{rP{1cm}P{1.5cm}P{1cm}P{1cm}} 
\hline
     & \vspace{-0.2em}\textbf{Null}      & \textbf{Error-amplifying}   & \vspace{-0.2em}\textbf{Guidance}  & \vspace{-0.2em}\textbf{All}          \\ 
\hline
\rowcolor[rgb]{0.753,0.753,0.753} \textbf{Number Per Group}  & 12                   & 13                   & 13                   & 38                    \\
\textbf{Gender}                                              &                      &                      &                      &                       \\
Male                                                         & 3                    & 6                    & 3                    & 13                    \\
Female                                                       & 6                    & 7                    & 10                   & 25                    \\
\textbf{Healthcare Trainee}                                  &                      &                      &                      &                       \\
Yes                                                          & 4                    & 7                    & 7                    & 19                    \\
No                                                           & 8                    & 6                    & 6                    & 19                    \\
\textbf{Handedness}                                          &                      &                      &                      &                       \\
Right                                                        & 11                   & 12                   & 12                   & 35                    \\
Left                                                         & 1                    & 1                    & 1                    & 3                     \\
\textbf{Exposure to dVRK}                                    &                      &                      &                      &                       \\
Yes                                                          & 1                    & 1                    & 1                    & 3                     \\
No                                                           & 11                   & 12                   & 12                   & 35                    \\
\textbf{Exposure to dVSS}                                    &                      &                      &                      &                       \\
Yes                                                          & 1                    & 1                    & 1                    & 3                     \\
No                                                           & 11                   & 12                   & 12                   & 35                    \\ 
\hline
                                                             & \multicolumn{1}{l}{} & \multicolumn{1}{l}{} & \multicolumn{1}{l}{} & \multicolumn{1}{l}{} 
\end{tabular}}
\label{demographictable}
\end{table}

Institutional Review Board approval (\#22514) was granted to recruit novice users of the da Vinci Surgical system, including both healthcare and non-medical trainees from Stanford University. Demographics are given in Table \ref{demographictable}. Forty surgical novices were recruited by e-mail, consented, and enrolled in the study. Surgical novices were defined as participants who had spent less than an hour total in their lifetime on any da Vinci system, including any commercially available da Vinci system, the dVRK, or the da Vinci Skill Simulator (dVSS). Of the 40 total subjects, 20 healthcare trainees and 20 non-medical trainees were recruited. Healthcare trainees were medical students, physician assistant students, and one resident. Non-healthcare trainees were graduate engineering students in the School of Engineering. Two subjects, one healthcare trainee and one non-medical trainee, left the study voluntarily after the first day. Data for these participants was excluded from analysis. 

Participants were told they would be required to participate over five consecutive days and asked about their experience using da Vinci robotic systems prior to the first day of participation. All trainees were considered RMIS novices as defined by our criteria. On the first day of the study, participants learned the details of study participation and gave informed consent. Participants were then shown a video of expert performance (see supplementary material) and instructed on how to use the dVRK. Participants were only instructed on the use of the teleoperation activation pedal and were not allowed to use the clutch to change the position mapping between the surgeon-side grippers and virtual manipulators. The virtual environment remained fixed such that the camera view and zoom level could not be altered. Participants were told that they would be evaluated on accuracy of movement and time. They were also told to ask for a reset if the ring were to fall to the floor of the environment. Following orientation, participants completed five trials in the null (no force) field to be used as a baseline of performance. Participants were subsequently randomized to one of three training groups: guidance force field, error-amplifying force field, and null field. For the remainder of the first study day, participants completed 15 trials in their assigned training field. On study days two through four, participants completed 20 trials in their assigned training field. On the final study day, participants completed 20 trials in the null field and filled out a post-survey. 

\subsection{Performance Metrics}

\begin{figure}[t]
\centering\includegraphics[width=\linewidth]{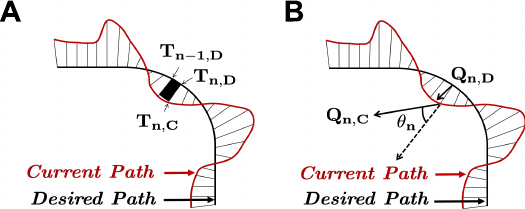}
\caption{(A) Desired ($T_{n,D}$), preceding desired ($T_{n-1,D}$), and current ($T_{n,C}$) ring positions used to calculate Translational Path Error, Rotational Path Error, and Combined Error-Time. (B) Desired ($Q_{n,D}$) and current ($Q_{n,C}$) ring orientations used to calculate Rotational Path Error and Combined Error-Time.}
\label{perfmetrics}
\end{figure}

\hlight{
User performance was evaluated using trial time, translational path error, rotational path error, and combined error-time. Trial time was calculated as the time in seconds elapsed from when the ring passed the starting point of the desired path to when the ring passed the end point of the desired path. This metric quantifies the speed of task completion, which is important for surgical proficiency. As in \cite{coad2017training}, Translational Path Error (TPE) was calculated by summing the products of the translational distance error between the current ($T_{n,C}$) and desired ($T_{n,D}$) ring positions and the distance between the desired ring position ($T_{n,D}$) and its preceding value ($T_{n-1,D}$) at the $n^{\text{th}}$ time step over the entire duration of the trial and is described as
\begin{equation}
\label{TPE}
    \textrm{TPE} = \sum_{n=1}^{N} \textrm{Dist}[T_{n,D} , T_{n,C}]\, \textrm{Dist}[T_{n,D} , T_{n-1,D}],
\end{equation}
where $n$ is the discrete time step, $N$ is the total number of discrete time steps in the entire trial, $\textrm{Dist}[T_1, T_2] = \sqrt{(X_2 - X_1)^2 + (Y_2 - Y_1)^2+ (Z_2 - Z_1)^2}$ for a given $ T_1= (X_1,\ Y_1,\ Z_1)$, $T_2= (X_2,\ Y_2,\ Z_2)$. Likewise, for the Rotational Path Error (RPE), the ring angular distance error between the current ($Q_{n,C}$) and desired ($Q_{n,D}$) ring orientations was used instead of the translational distance error to obtain the angular equivalent of the TPE such that
\begin{equation}
    \label{RPE}
    \textrm{RPE} = \sum_{n=1}^{N} \theta_{n}\, \textrm{Dist}[T_{n,D} , T_{n-1,D}],
\end{equation}\vspace{0.5em}
where $\theta_{n}$ = $H_{aa}$ $[Q_{n,D}Q_{n,C}^{-1}]$ such that $H_{aa}$ $[Q_2 \ Q_1^{-1}]$ represents the axis angle transformation of the quaternion that rotates an initial orientation $Q_1$ into a final orientation $Q_2$. The quantities used to compute the TPE and RPE are described graphically in Figure \ref{perfmetrics}.}

\hlight{ As in \cite{coad2017training}, Combined Error-Time (CET) was calculated as a combination of Trial Time, TPE, and RPE using the following equation:
\begin{equation}
    \label{CET}
    \textrm{CET}= \textrm{Time} \times (\textrm{RPE} + c \ \textrm{TPE})
\end{equation}
where $c$ is a constant factor of 17.27 rad/mm derived from the ratio of the average rotational path error and the average translational path error across all participants and all baseline and final trials. Evaluation metrics were computed by averaging the last five trials on the last day. Baseline metrics were computed by averaging the first five trials on the first day which were performed under no force feedback. Improvement was calculated for each metric as the difference between the final day evaluation and the baseline.}

\subsection{Statistical Analyses}

Metrics were non-normally distributed between groups, thus we used non-parametric statistical tests to evaluate differences in performance. Kruskal-Wallis (KW) tests were performed at baseline and on final evaluation with the different metrics as dependent variables and the training group as the independent factor. Post-hoc testing was then performed using the Dunn Test (DT) with Holm adjustment to determine significant differences between pairs of groups. Pairwise Wilcoxon signed-rank tests were used to determine significance of within-participants improvement from baseline to final evaluation. 

The KW test allows for hypothesis tests on non-normally distributed data, but is limited to analysis of a single factor. To investigate the influence of baseline skill on final evaluation performance for the different training conditions, we use linear regression with log-transformed metrics. The log transformation results in normally distributed data that can be used for linear regression but has the tradeoff of being less interpretable. The model is specified in statistical notation as
\begin{equation}
    \label{eqn:linear_model}
    \log(m_{e}) \sim  \log(m_{b}) + \kappa + \log(m_{b}) \times \kappa 
\end{equation}
where $m_e$ is the mean of the metric during the evaluation phase, $m_b$ is the mean of the metric during the baseline phase, and $\kappa$ is the force training condition, either guidance, error-amplifying, or null, as a categorical factor.

Hypothesis tests were conducted on the linear regression using F-tests to compute the statistical significance of each of the terms in the model. When a significant interaction between the log of the baseline metric and training condition was found, pairwise comparisons of the interaction between conditions with Bonferroni correction were performed.

\section{Results}

\subsection{Training Data}

\begin{figure*}[!tp]
 \centering
 \includegraphics[width=\linewidth]{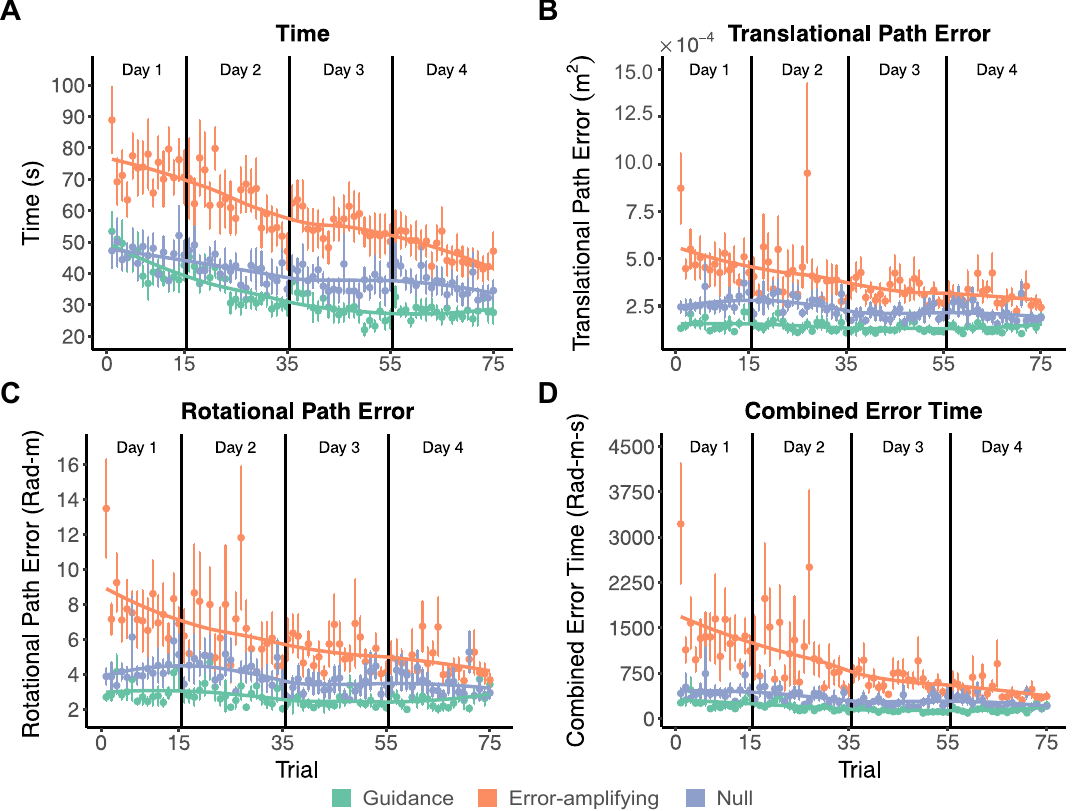}
 \caption{Learning curves for each of the four metrics over the four training days. Vertical lines delineate separation between training days. Trend lines were calculated using locally estimated scatterplot smoothing (LOESS). Error bars represent the standard error of the mean value of a metric for a particular group on a particular trial. Subplots represent learning curves for (A) time to completion, (B) translational path error (TPE), (C) rotational path error (RPE), and (D) combined error-time (CET). }
 \label{TrainingLCs}
 \end{figure*}
 
\begin{figure*}[!tp]
\centering
\includegraphics[width =\linewidth]{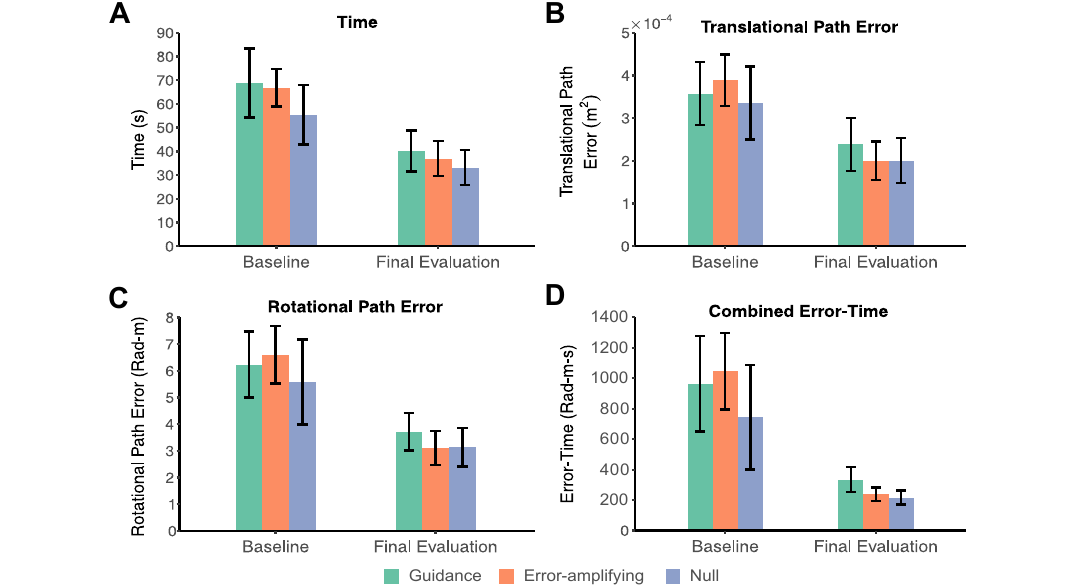}
\caption{Comparison of performance metrics for each group at baseline and on the final day of evaluation. Error bars represent the 95\% C.I. Subplots represent mean baseline and final evaluation values for each group in (A) time to completion, (B) translational path error (TPE), (C) rotational path error (RPE), and (D) combined error-time (CET).}
\label{RawValuesAggregate}
\end{figure*}

 \hlight{Learning curves for all metrics over the four training days are plotted in Figure \ref{TrainingLCs}. Due to a hardware fault, one of the participants assigned to the guidance force field only managed to train for thirteen out of the assigned twenty trials on day three. This data was excluded from Figure \ref{TrainingLCs}.}
 
 Participants assigned to the guidance field showed superior performance within their condition on all metrics throughout all four days of training compared to those who trained in the null and error-amplifying fields. Given the assistive nature of the guidance field, this was an expected result. In contrast, error-amplifying field participants showed inferior performance on all four days of training in all metrics. Given the inhibiting nature of the error-amplifying field which provided an additional challenge to participants, this result was also expected. Guidance field participants showed the least amount of improvement throughout the four days of training, while error-amplifying field participants had the largest improvements in training over the four days. There was a significant difference between groups in all metrics (KW; $p \leq$ 0.01 for all metrics) on the first training day, with error-amplifying field participants having significantly worse performance compared to the null group and guidance group. This group continued to have significantly worse performance within their training condition compared to the other two groups on the last day of training (KW; $p \leq$ 0.01). Ultimately, these results show that a participant's training condition significantly affected behavior during training. \\
 

 
 \subsection{Baseline vs. Final Performance: Aggregate Analysis}
 


\begin{figure*}[!tp]
\centering
\includegraphics[width=\linewidth]{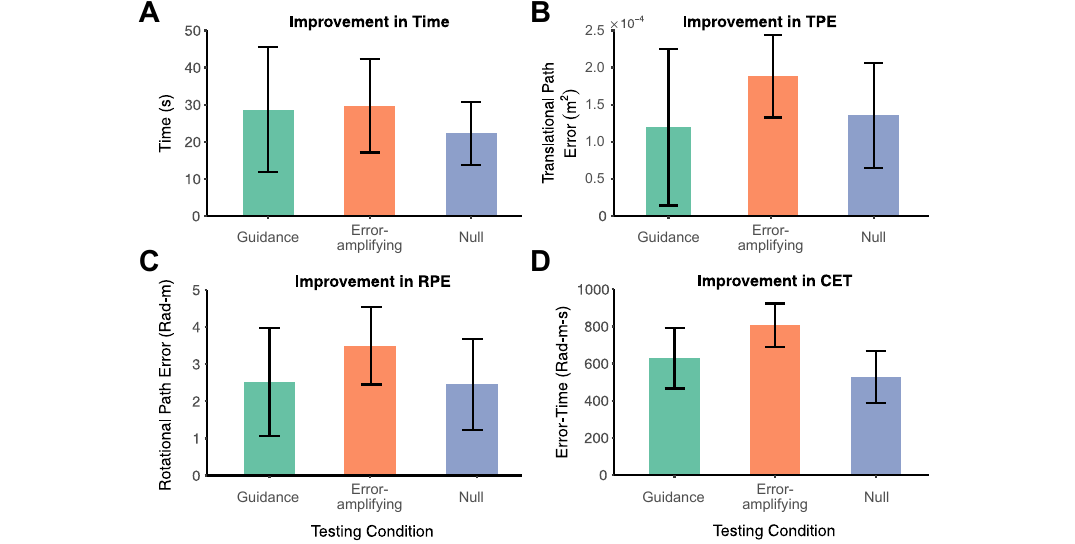}
\caption{Improvement of each group within each metric. Error bars represent the 95\% C.I. Subplots represent mean improvement from baseline to final evaluation in (A) time to completion, (B) translational path error (TPE), (C) rotational path error (RPE), and (D) combined error-time (CET).}
\label{ImprovementMetrics}
\end{figure*}

Figure \ref{RawValuesAggregate} compares the trial averaged performance for each group at baseline and on the final day of evaluation. This is also quantified by the trial averaged amount of improvement for each performance metric as shown in Figure \ref{ImprovementMetrics}. There was no significant difference found between groups for all metrics: time (KW; $\chi^2$ = 5.958, df = 2, $p =$ 0.0508), TPE (KW; $\chi^2$ = 1.071, df = 2, $p =$ 0.5850), RPE (KW; $\chi^2$ = 1.805, df = 2, $p =$ 0.4060), and CET at baseline (KW; $\chi^2$ = 3.922, df = 2, $p =$ 0.1410).
 
The groups significantly differed in CET on the final day (KW; $\chi^2$ = 6.932, df = 2, $p =$ 0.0312). Further analysis revealed a significant difference between guidance and null groups (DT; $Z =$ -2.531, $p =$ 0.0341). Of the three groups, the error-amplifying field participants had the lowest CET value on final evaluation, though this was not significant. All groups significantly improved their intragroup performance from baseline. 

For all metrics, the training condition was not found to be a significant predictor of improvement scores. Compared to the null field and guidance field groups, the error-amplifying field participants experienced the most improvement in all metrics (Figure \ref{ImprovementMetrics}). Guidance field participants experienced the least improvement from baseline of the three groups in TPE, RPE and CET (Figure \ref{ImprovementMetrics}B, C and D).

 \subsection{Influence of Baseline Ability and Training Condition on Learning}

\begin{figure*}[!htp]
\centering
\includegraphics[width=0.9\linewidth]{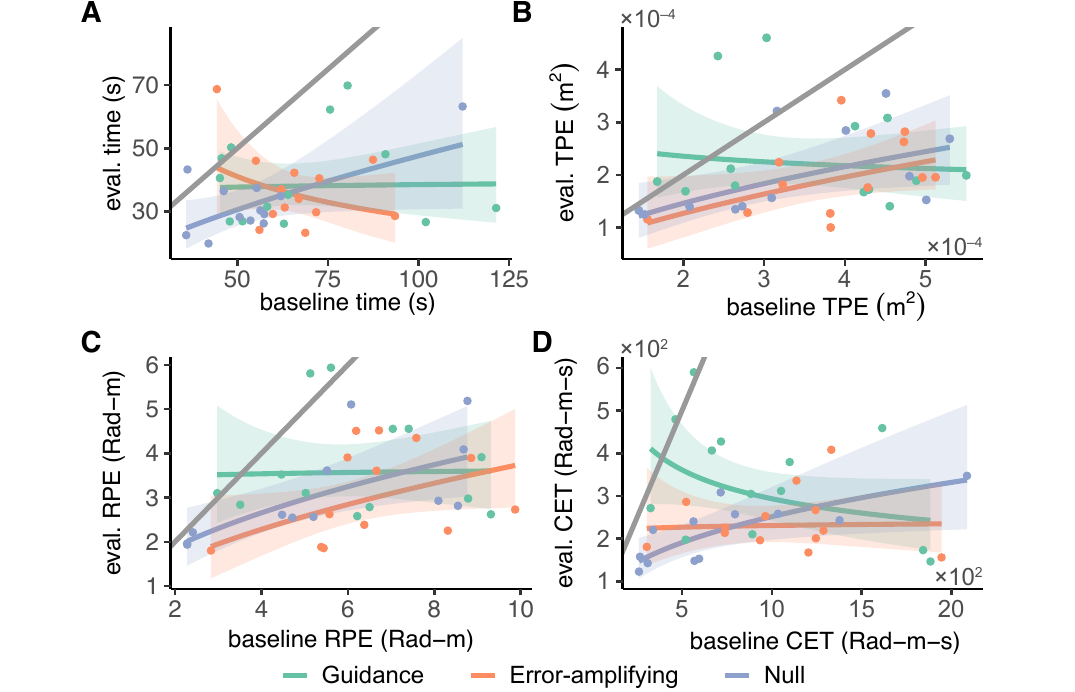}
\caption{Plots of linear regression fits of the log mean evaluation performance to the log mean baseline performance for (A) time, (B) translational path error, (C) rotational path error, and (D) combined error-time transformed back to the linear scale. Individual data points are for each subject. Shaded error bands denote 95\% C.I. The unity line (gray) separates the region where the evaluation performance was better than the baseline and the region where it was worse. Being below the unity line indicates improvement over the baseline.}
\label{linear_model}
\end{figure*}

 Figure \ref{linear_model} compares the trial averaged final day evaluation performance to the trial averaged baseline performance for each subject and presents the regression fits, which were performed with the log transformed metrics and transformed back to the linear scale, for each training condition. For time to completion, there were no significant effects of training condition and baseline time to completion found. For participants who trained in the error-amplifying condition, those with higher baseline time to completion showed lower time to completion in evaluation (Figure \ref{linear_model}A). In contrast, for participants who trained in the null condition, those with higher baseline time to completion showed higher time to completion in evaluation. For participants who trained in the guidance condition, little to no relationship was found between baseline time to completion and time to completion in evaluation. 

The effect of baseline TPE and RPE on their respective metrics in evaluation was statistically significant  (TPE \textendash \, F(1,32) = 4.97, p = 0.033; RPE \textendash  \, F(1,32) = 11.44, p = 0.002). For the error-amplifying and null conditions, a higher baseline TPE resulted in higher TPE in evaluation (Figure \ref{linear_model}B). The opposite trend was found for the guidance condition, such that the linear regression predicted a decrease in TPE on final evaluation for an increase in baseline TPE. For all training conditions, a higher baseline RPE resulted in higher RPE in evaluation (Figure \ref{linear_model}C). In the original metric space for RPE, there is a diminishing rate of increase in RPE in evaluation as baseline RPE increases.

 For CET, there was a significant interaction between training condition and baseline CET on predicting CET in evaluation ($F(2,32) =$ 4.81, $p =$ 0.015). There was also a significant effect of condition on CET in evaluation ($F(2,32) =$ 4.80, $p =$ 0.015). Post-hoc tests found a significant difference in the change in CET in evaluation with respect to baseline CET between the guidance and the null training conditions ($t(32) =$ -3.084, $p =$ 0.013). The linear regression predicts a decrease in CET in evaluation for an increase in baseline CET for the guidance training condition as seen in Figure \ref{linear_model}D. The trend for the null training condition was the opposite. In the original metric space, there is a diminishing rate of decrease in CET in evaluation for increasing baseline CET in the guidance condition, and a diminishing rate of increase in CET in evaluation for increasing baseline CET in the null condition. The linear regression predicted little difference in CET in evaluation over the range of experimentally observed baseline CETs for the error-amplifying condition.






\section{Discussion}

\subsection{Training Data}
The error-amplifying field participants showed the most improvement within their training environment from the first day to the last day of training in all metrics compared to the null and guidance field groups. Despite being in a more challenging environment, the error-amplifying field participants showed similar performance to the null field and guidance field participants by the end of the four training days. While the slope of each learning curve for the error-amplifying field participants had a visually flatter slope on the fourth day compared to the previous three days, the slope notably had not fully plateaued by the end of the final training day (Figure \ref{TrainingLCs}). This suggests that the error-amplifying field group may have still been learning on the final day, and thus may have surpassed the other groups in performance metrics if the training period had been extended.

\subsection{Baseline vs. Final Performance}
Error-amplifying field participants also had the most improvement from baseline to final evaluation in all metrics. While the difference in improvement between groups was not statistically significant, this trend in performance was especially promising considering the relative differences in baseline performances. Most notably, the error-amplifying field group was significantly slower to complete the task than the null group at baseline, but no significant difference in time to completion was found on final evaluation.

An important finding in the study was the significant difference found in CET between groups on the final day. This metric was used to evaluate performance in the context of the speed-accuracy tradeoff. In surgery, the balance between speed and accuracy is especially important; it is appropriate to move quickly when the stakes of a surgical step are low (e.g. sweeping avascular structures aside), versus moving slowly and accurately during the steps where an error could be catastrophic (e.g. dissection around major blood vessels). Most notable for this metric is the significantly inferior performance of the guidance field group. This group had by far the best performance throughout the four training days, but showed less improvement in performance compared to the error-amplifying field group and had by far the worst overall performance on the final evaluation. The superior performance throughout training but unreplicated improvement in performance on final evaluation further support the notion introduced in Schmidt et al. \cite{schmidt1989guidancehypothesis}, and Powell and O'Malley \cite{powell2012sharedcontrol}, that added assistance may be beneficial in surgery if it is able to be applied intra-operatively, but can lead to a dependence on the assistive forces which hinder performance when the assistive forces are removed.

\subsection{Influence of Baseline Ability and Training Condition on
Learning}
The analysis of the influence of baseline performance on performance in evaluation suggests that better baseline ability in reducing path errors predicts better ability to do so after training. With respect to translational and rotational path errors, users of high initial ability (lower path errors at baseline) could benefit from the difficulty of the task under the null or error-amplifying field. They achieved better performance after training than those with lower initial ability within their training condition. The benefit while training with the guidance force field was more limited, with users of all initial abilities achieving similar performance after training. This suggests that users of high initial ability benefited less from haptic guidance compared to those with lower skill within their training condition. 

The interaction between baseline time to completion and training condition was not found to be significant in predicting evaluation time to completion. However, Figure \ref{linear_model}A does show evidence of crossed slopes, with the error-amplifying training condition seemingly benefiting initially slower (high baseline time) participants more than initially faster participants within the same training condition. This effect is likely due to the destabilizing nature of the error-amplifying force field, which potentially rewards slower movements during training that translate to greater stability and therefore greater speed when inhibiting forces are removed on final evaluation.

The significant interaction between training condition and baseline CET in predicting evaluation CET suggests that the type of force training differently affects people with different baseline ability levels to trade off speed and accuracy. Participants who initially had a lower baseline CET appeared to benefit from the moderate level of difficulty of the null training condition. Within this group, these participants had a lower CET in evaluation compared to those who had higher baseline CET. The logarithmic trend in Figure \ref{linear_model}D highlights that the benefit of the null field training for higher skilled participants is exponentially greater than the benefit to lower skilled participants. 

The opposite was found for the guidance group, where users who had low baseline CET performed worse in evaluation than those who had high baseline CET. In the guidance group, the detrimental effects on those with low baseline CET are exponentially higher, as shown in Figure \ref{linear_model}D. The observed trend implies that those with higher baseline skill (lower baseline CET) rely more on the guidance to aid task performance in contrast to those with lower baseline skill (higher baseline CET), who might rely more on guidance to aid with learning the task. 

The trend of CET in evaluation against baseline CET for the error-amplifying field group showed that, regardless of initial ability, all participants reached similar levels of CET in evaluation. This is an unexpected result in that the difficulty of the error-amplifying field would be greatest and thus provide a more optimal task difficulty for higher ability participants to leverage for learning compared to lower ability participants. One possible reason for this unexpected result might be that despite the higher difficulty of the error-amplifying training condition, the alteration of the task dynamics by the rendered forces resulted in participants learning how to perform the task with modified physics that conferred no benefit in performance in the task with the correct physics. Such a limitation of haptic rendering was hypothesized by Powell and and O'Malley for guidance and could plausibly be extended to error amplification \cite{powell2012sharedcontrol}. 

\subsection{Future Work}
While we observed meaningful performance differences among participants trained under different conditions, the simplicity of the task could have limited the magnitude of these differences. While the movements in each of the three Cartesian coordinates were initially challenging for participants, it is possible that participants with innately high ability were able to master the task in a shorter amount of time than that allotted for the study. This may have accounted for the difference in time to completion between null participants and those in the guidance field and error-amplifying field groups at baseline. Conversely, it is also possible that more training days were necessary to observe a difference in performance between the three training groups. While the learning curves in Figure \ref{TrainingLCs} appear to flatten for the guidance and null groups by training days 3 and 4, the learning curves for the error-amplifying field group notably did not flatten by the end of training. This observation suggests that the error-amplifying field group was still actively learning at the end of the training period, which further suggests that these participants may have had further improvement in their performance given more time to train. Given this observation, coupled with the similarity in performance between the null field and error-amplifying field groups in evaluation, the error-amplifying field group may have surpassed the null group with more days of training added to the training period. 

Another factor that could have diminished the amount of learning and performance differentiation among groups is the nature of the provided visual feedback. In this study, the color of the ring changed from red to yellow as the distance of its center to the closet point on the wire decreased. Up to a certain distance away from the wire, the shift in color towards full yellow might be imperceptible to participants. Thus, if they relied heavily on this visual feedback cue, they might have aimed to keep the ring close enough to the wire so that it was perceived as yellow without further perfecting their positional accuracy. Furthermore, orientation error was not visually communicated by color due to the confusion it caused in pilot testing when combined with position error color feedback, hence participants might not have aimed to reduce orientation error as much as position error. In the guidance field and error-amplifying field groups, this effect would be less pronounced due to the availability of force feedback. However, it could have been influential for the groups training in the null condition.



Apart from addressing the above limitations of the current experiment, future iterations of this experiment could evaluate performance in the null field at the end of each training session.  This evaluation would allow for creation of a learning curve in the baseline environment across the four days. Another possible future implementation is to have participants train in the guidance field to gain comfort with the task, followed by training in the error-amplifying field to gain expertise and reduce the cognitive load of the task in the baseline environment. 

Beyond this particular virtual environment, a more applied future direction of this work would be to translate the use of error amplification and assistive/inhibiting force fields to more realistic surgical tasks, such as suturing, knot tying, stapling, or vessel ligation. Relatedly, while modern technology allows for implementation of realistic physics within virtual environments, some element of the true physical properties of materials is lost in virtual reality. For that reason, a study with teleoperation in a physical environment that implements error amplification would be more realistic and translatable to clinical medicine.  

\section{Conclusions}

In this work we investigated how surgical novices controlling a minimally invasive surgical robot in a virtual reality ring-on-wire task performed after training under error-amplifying forces. We compared their performance to those who trained under guidance forces or with no force feedback. Performance was measured using orientation and positional accuracy metrics, task completion time, and a composite error-time metric that quantified an individual's speed-accuracy tradeoff.

Participants who trained to complete the ring-on-wire task with the aid of assistive error reducing forces had superior performance throughout the training period but inferior performance at final evaluation in the baseline, non-assisted environment. Participants who trained to complete the ring-on-wire task with the addition of inhibiting, error amplifying forces showed the most improvement in performance from their baseline. The baseline ability of the participant affected the amount of benefit that could be derived from each force training condition. For path error, participants with high baseline path error benefited less from force guidance than those with lower error. In terms of the ability to trade off speed and accuracy, training with no assistive or inhibitory forces benefited those with high ability more than those with low ability. The opposite effect was seen in those who trained with guidance \textendash\,training with assistive or inhibitory forces most benefited those with low baseline ability. Participants who trained in the divergent field did not have a plateau in performance after four days of training, and we hypothesize that with more days to train, divergent field participants would have superior performance. 


Our results indicate that training with error amplifying forces can be useful for developing skill in RMIS. Specifically, trainees with high initial levels of path accuracy or slow speed may stand to see improvement in those respective skill dimensions. This insight provides a step towards designing more personalized RMIS haptic training systems that provide learner-appropriate feedback to optimize trainee skill acquisition.

\section*{Acknowledgements}

The authors would like to acknowledge the members of the Collaborative Haptics and Robotics in Medicine (CHARM) Laboratory at Stanford University for their support in the design and implementation of this project. We would also like to thank Elyse Chase who aided in the early development of this project. 

\bibliographystyle{IEEEtran}
\bibliography{IEEEabrv,sample.bib}

\section{Biography Section}
 
\vspace{-2\baselineskip}
\begin{IEEEbiography}[]{Yousi Oquendo} received the M.S.E. degree in robotics engineering from the University of Pennsylvania, Philadelphia, PA and the M.D. degree from Stanford University School of Medicine, Stanford, CA in 2016 and 2022, respectively. She is currently an Orthopaedic Surgery Resident Physician at the Hospital for Special Surgery, New York, NY. Her research interests include the measurement and assessment of surgical skill in minimally invasive surgery.
\end{IEEEbiography}
\vskip -2\baselineskip plus -1fil
\begin{IEEEbiography}
[]{Margaret Coad}(Member, IEEE) received the M.S. and Ph.D. degree from Stanford University in 2017 and 2021, respectively, all in mechanical engineering. She is currently an Assistant Professor at the University of Notre Dame in the Department of Aerospace and Mechanical Engineering. Her research interests include the design, modeling, and control of innovative robotic systems to improve human health, safety, and productivity.
\end{IEEEbiography}
\vskip -1\baselineskip plus -1fil
\begin{IEEEbiography}
[]{Sherry Wren} received the M.D. degree from Loyola University Chicago Stritch School of Medicine, Chicago, IL in 1986. She is currently a Professor, Vice Chair, and Director of Global Surgery at Stanford University, Stanford, CA as well as Clinical Surgery Director at the Palo Alto Veterans Hospital, Palo Alto, CA. Her clinical and research interests include gastrointestinal cancer treatment and surgical robotics.
\end{IEEEbiography}
\begin{IEEEbiography}[]{Thomas Lendvay} 
received the M.D. degree from Temple University Lewis Katz School of Medicine, Philadelphia, PA in 1999. He is a Professor of Urology at the University of Washington School of Medicine, Seattle, WA, and the co-founder of three medical technology companies, C-SATS, Singletto, and Tend-Health Inc., all in Seattle, WA. His research interests include surgical technology, medical device design, and development of scalable technologies to enhance healthcare. 
\end{IEEEbiography}
\vskip -2\baselineskip plus -1fil
\begin{IEEEbiography}
[]{Ilana Nisky} (Senior Member, IEEE) received the B.Sc. (summa cum laude), M.Sc. (summa cum laude), and Ph.D. in biomedical engineering from Ben-Gurion University of the Negev, Beersheba, Israel, in 2006, 2009, and 2011, respectively. She is currently a Professor with the Department of Biomedical Engineering, Ben-Gurion University of the Negev. Her research interests include human motor control, haptics, robotics, human and machine learning, teleoperation, and robot-assisted surgery. 
\end{IEEEbiography}
\vskip -2\baselineskip plus -1fil
\begin{IEEEbiography}[]{Anthony Jarc}received the M.S. and Ph.D. degrees from Northwestern University, Evanston, IL in biomedical engineering in 2010 and 2011, respectively. He is currently the Senior Director of Data and Analytics at Intuitive Surgical Inc., Sunnyvale, CA. His research interests include machine learning-enabled digital solutions for improved robotic-assisted surgery and surgical data science.
\end{IEEEbiography}
\vskip -2\baselineskip plus -1fil
\begin{IEEEbiography}[]{Allison Okamura}(Fellow, IEEE) received the B.S. degree from the University of California, Berkeley, in 1994, and the M.S. and Ph.D. degrees from Stanford University, Stanford, CA, in 1996 and 2000, respectively, all in mechanical engineering. She is currently the Richard W. Weiland Professor in the School of Engineering and Professor of Mechanical Engineering at Stanford University, Stanford, CA. Her research interests include haptics, teleoperation, medical robotics, virtual environments and simulation, neuromechanics and rehabilitation, prosthetics, and engineering education.
\end{IEEEbiography}
\vskip -2\baselineskip plus -1fil
\begin{IEEEbiography}[]{Zonghe Chua}(Member, IEEE) received the M.S. and Ph.D. degrees from Stanford University, Stanford, CA, in 2020 and 2022, respectively, all in mechanical engineering. He is currently an Assistant Professor of Electrical Engineering at Case Western Reserve University, Cleveland, OH. His research interests include teleoperation, robot-assisted surgery, haptic feedback, and machine learning approaches for augmented human-robot interfaces.
\end{IEEEbiography}



\vfill

\end{document}